\documentclass[conference]{IEEEtran}
\IEEEoverridecommandlockouts
\usepackage{cite}
\usepackage{comment}
\usepackage{amsmath,amssymb,amsfonts}
\usepackage{algorithmic}
\usepackage{graphicx}
\usepackage{textcomp}
\usepackage{todonotes}
\usepackage{xcolor}
\usepackage{url, hyperref}
\def\BibTeX{{\rm B\kern-.05em{\sc i\kern-.025em b}\kern-.08em
    T\kern-.1667em\lower.7ex\hbox{E}\kern-.125emX}}

\newcommand{\pb}[1] {{\color{red}{#1}}}

\usepackage[switch]{lineno}

\usepackage{authblk}

\begin{document}

\title{Sequence Length Scaling in Vision Transformers \\ for Scientific Images on Frontier}

\author[1]{Aristeidis Tsaris \textsuperscript{*}}
\author[2]{Chengming Zhang}
\author[1]{Xiao Wang}
\author[1]{Junqi Yin}
\author[1]{Siyan Liu}
\author[1]{Moetasim Ashfaq}
\author[1]{Ming Fan}
\author[1]{\\Jong Youl Choi}
\author[3]{Mohamed Wahib}
\author[1]{Dan Lu}
\author[1]{Prasanna Balaprakash}
\author[1]{Feiyi Wang}
\affil[1]{Oak Ridge National Laboratory, Oak Ridge, United States}
\affil[2]{Indiana University Bloomington, Bloomington, United States}
\affil[3]{RIKEN Center for Computational Science, Kobe, Japan}

\maketitle

\begingroup\renewcommand\thefootnote{*}
\footnotetext{Corresponding author: tsarisa@ornl.gov}
\endgroup

\begin{abstract}

Vision Transformers (ViTs) are pivotal for foundational models in scientific imagery, including Earth science applications, due to their capability to process large sequence lengths. While  transformers for text has inspired scaling sequence lengths in ViTs, yet adapting these for ViTs introduces unique challenges. We develop distributed sequence parallelism for ViTs, enabling them to handle up to 1M tokens. Our approach, leveraging DeepSpeed-Ulysses and Long-Sequence-Segmentation with model sharding, is the first to apply sequence parallelism in ViT training, achieving a 94\% batch scaling efficiency on 2,048 AMD-MI250X GPUs. Evaluating sequence parallelism in ViTs, particularly in models up to 10B parameters, highlighted substantial bottlenecks. We countered these with hybrid sequence, pipeline, tensor parallelism, and flash attention strategies, to scale beyond single GPU memory limits. Our method significantly enhances climate modeling accuracy by 20\% in temperature predictions, marking the first training of a transformer model on a full-attention matrix over 188K sequence length.

\end{abstract}


\section{Introduction} \label{sec:1}

In the rapidly evolving landscape of artificial intelligence and machine learning (AI and ML), Vision Transformers (ViTs) have become notable for their ability to model long-range dependencies between tokens in time-series and imaging data \cite{Khan_2022}. ViTs are proving effective in building foundational models for scientific imagery applications, including research in Earth system science and geospatial analysis \cite{chen2022scaling, xiong2024all, cha2023billionscale}. Additionally, ViTs are commonly used as the architecture of choice for image encoders in multimodal foundation models, highlighting their importance in processing complex image inputs \cite{li2023multimodal}. 
Despite their crucial role in foundation model (FM) development, ViTs have not scaled to the same extent as Large Language Models (LLMs). To date, LLMs have expanded to encompass over one trillion parameters~\cite{10.5555/3586589.3586709}, whereas the ViTs have \textit{only} reached a pinnacle with 22 billion parameter models, specifically the ViT-22B \cite{dehghani2023scaling}.

ViTs face distinct challenges compared to the transformers used in LLMs, such as GPTs. The architectural differences between these models significantly influence their operation, training, and application. Transformers in LLMs are primarily decoder-only models focused on processing and generating textual data. Their input consists of embeddings of tokenized text, which operates within a well-defined predictive space limited by the vocabulary size. This bounded space simplifies the task of predicting the next word in a sequence, allowing for efficient scaling primarily through increases in model size. Conversely, ViTs embody an encoder-decoder architecture, adding complexity layers to their functionality. Unlike LLMs, ViTs are trained on image and video data converted into a sequence of tokens derived from patches. This method of data representation is inherently more complex than the tokenized text embeddings used by LLMs. The visual domain does not benefit from a constrained predictive space like language. Instead, the space of possible future images or visual outputs is virtually limitless, presenting a significant challenge for ViTs when forecasting or generating new visual content.
This fundamental distinction between the textual versus visual data necessitates different scaling strategies. For LLMs, enhancing model size has been a focal point for improving performance. However, for ViTs, sequence length distribution emerges as a crucial aspect for further scaling. The infinite possibilities in visual representation demand that ViTs process larger sequences of visual data. Moreover, the encoder-decoder framework of ViTs must efficiently translate raw visual inputs into a sequence of tokens and then interpret these tokens across a broad range of tasks. The challenge is compounded by the need for ViTs to handle these tasks with a high degree of accuracy and versatility, given the unbounded nature of visual data.



For LLMs and ViTs, extending sequence lengths has been identified as a method to enhance model performance. While conventional training for ViTs on natural images typically employs a sequence length of 256, even for expansive models~\cite{dehghani2023scaling}, scientific foundational models (FMs) such as those used in ClimaX \cite{nguyen2023climax}, a foundation model for weather and climate forecasting, leverage significantly longer sequences, up to 8K or more. Scientific imagery, including microscopy or remote sensing data, often surpasses the resolution of standard images, reaching dimensions of 100,000 x 100,000 pixels~\cite{KIM2021101854,article-white,article-bc}. This high resolution, combined with the unique characteristics of scientific data, such as integrating observational data with simulation data, necessitates much longer sequence lengths for effective model training. For instance, training on the native resolution of ERA5 reanalysis data \cite{ERA5} with the ClimaX foundation model would require a sequence length of 277K, vastly exceeding the lengths used for images from consumer devices. Moreover, the complexity of scientific images is  compounded by their channel diversity. Unlike the correlated RGB channels of consumer device images, each channel in scientific images typically represents a distinct physical quantity, leading to a linear increase in the number of sequence lengths as the number of channels grows. This diversity and the need for handling spatiotemporal images or unstructured time-series data, such as video or scientific simulations, underscore the demand for longer sequence lengths in scientific applications. However, scaling sequence length is not without its challenges. The memory footprint of transformers scales quadratically with sequence length, posing significant constraints on training capacity with current hardware. As a result, training on longer sequences often requires trade-offs, utilizing sparse attention mechanisms or approximations of the full-attention matrix to manage computational demands.

Dense attention within transformer models facilitates a comprehensive interaction among all tokens in the input sequence, enabling the model to capture long-range dependencies and intricate patterns \cite{vaswani2023attention}. Contrastingly, sparse attention mechanisms \cite{wang2020linformer}, designed to mitigate the computational demands of dense attention, limit each token's interaction to a subset of other tokens. While this approach reduces computational complexity—often from quadratic to linear or sub-quadratic—it may not capture the full context as effectively as dense attention, particularly in tasks where the relationship between distant parts of the input is critical.

Justification for the use of dense attention despite its higher computational cost comes down to its unmatched ability to model complex interactions and dependencies across entire sequences. In scientific imagery, where the precise arrangement and relationship of pixels or data points can convey essential information, dense attention's comprehensive feature learning capability is invaluable. It ensures that no potential relationship or pattern is overlooked, a critical factor in accurately modeling and interpreting scientific imagery data. Moreover, dense attention's robustness in handling intricate data patterns makes it critical for advanced applications, including high-resolution image analysis, complex sequence prediction, and detailed textual interpretation, where the length of the context directly impacts model performance.

We developed an ultra-long vision transformer, with dense attention, that uses distributed sequence parallel methods for scientific images. Our approach is based on DeepSpeed-Ulysses \cite{jacobs2023deepspeed} and on Long Sequence Segmentation (LSS) \cite{wang2023ultralong} for distributing long sequences across GPUs as contiguous segments. DeepSpeed-Ulysses aims to optimize the distribution of computational load for self-attention across multiple GPUs, thereby enhancing the model's scalability and efficiency. LSS on the other hand addresses the direct dependency on self-attention. It employs fused communication and a double gradient averaging technique, reducing the complexity of aggregating partial self-attention scores. This approach aims to simplify the computational process and improve scalability by minimizing the communication overhead typically associated with dense attention in long sequences.

These two methods exhibit different communication patterns per attention layer. While DeepSpeed-Ulysses is expected to see quadratic growth in communication demand with an increase in parallel GPUs due to its distribution of self-attention, LSS aims for a more streamlined communication strategy, with only two calls per attention layer. We analyzed the scaling efficiency of each method separately, scaling the sequence length up to 1M sequence length, for model sizes up to 10B of parameters. 

Our analysis showed limitations and performance bottlenecks of each method, for example, DeepSpeed-Ulysses scaling size is limited by the number of attention heads of the model. And while flash-attention-v2 \cite{dao2023flashattention2} can help to fit very large sequences for small model sizes, for models that can't fit on a single GPU a hybrid parallelism is necessary. To that extent, we used Pipeline Parallelism (PP)~\cite{huang2019gpipe} with DeepSpeed-Ulysses and Tensor Parallelism (TP)~\cite{shoeybi2020megatronlm} with LSS so that we can train more efficient on longer sequence lengths compared to single parallel mode. 

There is a noted absence in the literature for ViT performance baselines, and with this work, we provide an empirical study for the practical applicability of each method in distributed environments at different scales. Also as top HPC centers in America and Europe are using AMD GPUs, with this work we are testing the AMD AI software stack on four state-of-the-art distributed methods, while showing how to best use the Frontier Supercomputer for large-scale AI on scientific images. We run at scale a 188K sequence length ViT model, while scaling the batch size up to 2,048 AMD MI250X GPUs measuring a 94\% scaling efficiency.

Additionally, we established the advantage of longer sequence training on a climate application by showing that we can get better accuracy by increasing the sequence length of the model. Specifically, we trained on the ERA5 climate dataset, and by increasing either the image resolution or the number of channels embedding to the model per time-step, we see a 20\% higher accuracy in temperature prediction. This is the first time a transformer model was trained with a full attention matrix on 188K sequence length.

In summary, our contributions are the following:

\begin{itemize}
\item We have developed a vision transformer capable of handling ultra-long sequence lengths through sequence parallelism. Our method leverages DeepSpeed-Ulysses and LSS parallel strategies, achieving sequence lengths of up to 1 million. This marks the first successful application of sequence parallelism in training Vision Transformers (ViTs), and by integrating model sharding, we attained a 94\% batch scaling efficiency across 2,048 AMD MI250X GPUs.

\item Our evaluation of sequence parallelism on ViTs covered models up to 10 billion parameters. The analysis identified significant performance bottlenecks, particularly in models with 5 and 10 billion parameters. To address these, we employed distributed sequence strategies in combination with Pipeline Parallelism (PP) and Tensor Parallelism (TP), as well as flash-attention-v2. This approach enabled further scaling of sequence lengths for models that exceed the memory capacity of a single GPU and included an empirical study across various scales.

\item We have showcased the advantages of utilizing extended sequence lengths in climate modeling, observing up to a 20\% increase in accuracy for temperature predictions. This achievement represents the first instance of successfully training a transformer model to process a full-attention matrix across such extensive sequence lengths, specifically 188K tokens.
\end{itemize}

The rest of the paper is organized as follows. In Sections~\ref{sec:bkg} and \ref{sec:related_work} we defined the concepts and the terms used in this paper, and discuss related work. Section \ref{sec:seq-vit}, shows the advantage of training on longer sequence lengths for a climate application, and we calculate the computation training need for scientific images on Frontier. Section~\ref{sec:vit-perf} shows scaling solutions and best approaches for different model sizes and sequence lengths. Finally, convergence runs and weak scaling performance are shown in Section \ref{sec:vit-all} with real climate data.

\section{Background} \label{sec:bkg}
This section describes the ViT model and the patch embedding methods used in this work, along with the various serial and distributed methods applied.  

\subsection{Patch Embedding and Sequence Length}
In most common cases, ViTs are applied to images captured by consumer devices, where the typical resolution is 100s pixels. To make a sequence of tokens for the encoder a patch size of 16x16 is typically selected. It is common practice for larger models to have smaller patch sizes, leading to a larger sequence of tokens \cite{dosovitskiy2021image, dehghani2023scaling}. General RGB images have three overlapping channels  on the same object, so 3D patches are usually chosen. For a 224x224x3 image using 3D patches of 16x16x3, we will end up with a sequence length of 192 tokens. 

On the other hand, scientific simulations, such as in Earth system science, or multi-sensor data acquisition systems, such as medical imaging, produce multi-dimensional data, and the channels are independent. In this case, 2D patches are more appropriate since the model needs to understand different physical quantities from sensors or variables. For example, for the same 224x224x3 image, if we apply 2D 16x16 patches we will end up with a three times larger sequence length. When we explicitly embed the full number of channels in the transformer, we will call this \textbf{Multi-Ch-ViT}.

Some approaches aggregate all the channels into one, as proposed by \cite{nguyen2023climax}, by including a multi-head attention layer in the encoder, i.e., the data query is a learnable vector. That way, the number of channels won’t increase the sequence length. We call this \textbf{Agg-Ch-ViT}.



\subsection{Data Parallel (DP)}
In data parallelism a copy of the entire model is replicated across each worker. Zero Redundancy Optimizer (ZeRO) \cite{ren2021zerooffload} removes the memory redundancies across data-parallel processes by partitioning the three model states (optimizer states, gradients, and parameters) across data-parallel processes instead of replicating them.


\subsection{Pipeline Parallel (PP)}
Pipeline parallelism~\cite{huang2019gpipe} can improve the efficiency and scalability of training large deep learning models, particularly those that are too large to fit into the memory of a single device or that require substantial computational power. Unlike data parallelism, which replicates the entire model across multiple devices and divides the dataset among them, pipeline parallelism divides the model itself into several smaller, sequential stages or segments. Each stage of the model is then placed on a different device.

\subsection{Tensor Parallel (TP)}
Tensor parallelism involves partitioning the model's tensors across different processors. Megatron-LM~\cite{shoeybi2020megatronlm} exploits the inherent structure of Transformer networks to implement tensor parallelism efficiently. It parallelizes both the Multi-Layer Perceptron (MLP) blocks and the self-attention blocks of a Transformer layer, using only a few synchronization primitives.



\subsection{Sequence Parallel (SP)}
Sequence parallelism, which distributes the self-attention computation in transformer models across the sequence dimension, has been an active area of research, with many developed methods employing various approaches. DeepSpeed-Ulysses~\cite{jacobs2023deepspeed}introduces a suite of system optimizations specifically designed for the efficient training of extreme long sequence transformer models. DeepSpeed-Ulysses partitions the input data along the sequence dimension, which allows the system to handle much longer sequences. For attention computation, it employs an all-to-all collective communication to ensure that each GPU receives a complete sequence, but only for a non-overlapping subset of the attention heads, allowing for parallel computation of attention across GPUs. 
LSS~\cite{wang2023ultralong} divides a long sequence into segments distributed among GPUs, with each GPU computing a partial self-attention for its segment. It also introduces a fused communication strategy to minimize the communication overhead. Dynamic Sequence Parallelism (DSP) \cite{zhao2024dsp} enables efficient sequence parallelism for multi-dimensional transformers by dynamically switching the parallelism dimension according to the current computation stage. Ring attention \cite{li2022sequence, liu2023ring} is another variant that organizes tokens in a ring structure, allowing each token to attend to a fixed number of adjacent tokens in the sequence. This method offers a compromise between dense and sparse attention, balancing computational efficiency with the ability to capture local and semi-distant relationships. However, ring attention may still fall short in tasks where long-range dependencies are crucial.






\subsection{Flash Attention v2}
FlashAttention~\cite{Dao2022FlashAttentionFA} significantly speeds up attention computation by tiling the attention matrix to reduce the number of memory reads and writes between the GPU's high bandwidth memory (HBM) and on-chip SRAM. 
FlashAttention-2~\cite{dao2023flashattention2} further accelerates attention computation by reducing the number of non-MatMult FLOPs and parallelizing computation in the sequence length dimension.

\section{Related Work} \label{sec:related_work}

As mentioned in Section \ref{sec:1}, literature reviews reveal an absence of references utilizing ViTs with sequence parallelism for baseline comparisons. The majority of existing studies concentrate on extending the sequence length within a single GPU to examine its impact on convergence rates. Machine learning datasets comprising natural images typically feature low resolutions, such as 224x224 pixels. A common approach to enlarging the receptive field involves reducing the patch size, thereby increasing the sequence length. Conversely, high-resolution datasets, notably those in remote sensing imagery, often resort to cropping images to 224x224 pixels \cite{xiong2024all, cha2023billionscale}. This practice significantly hampers the transformer's ability to apprehend the global context, thereby undermining the potential advantages of utilizing high-resolution images.


In the climate and weather domain, Vision Transformers (ViTs) and their variants have been increasingly adopted for rapid and accurate weather forecasting. The main objective of the ViT used on weather data is to predict from a weather snapshot $X_{t+}$ of a specific time $t$ the future weather forecast $X_{t+\Delta t}$ at lead-time  $\Delta t$.
The majority of the climate ViTs handle the long sequence complexity by replacing the dense attention matrix with some approximation. 
FourCastNet \cite{pathak2022fourcastnet} employs a Fourier transform-based token mixing scheme alongside a ViT backbone. This approach enables flexible and precise forecasts of global weather conditions ranging from 6 hours to 7 days ahead, covering high-resolution, rapid-timescale variables such as surface wind speed, precipitation, and atmospheric water vapor. Pangu-weather has introduced a 3D Earth-specific Transformer model that also leverages the ViT architecture. This model effectively captures the 3D atmospheric dynamics, delivering more accurate 7-day weather forecasts \cite{bi2023accurate} compared to traditional numerical weather prediction methods. Additionally, the FuXi model \cite{chen2023fuxi}, based on ViT, demonstrates performance on par with established forecasting methods, extending its forecasting capability to an impressive 15 days.


The above models, even though are trained on the ERA5 high-resolution dataset and use ViT as a backbone, they don't use the full attention matrix for the training. They mostly rely on approximations, or on sparse attention, such as Swin Transformer~\cite{liu2021Swin}, in order to fit that long sequence on a single GPU. On the other hand, Climax~\cite{nguyen2023climax} does use self-attention, but it is trained on lower resolution compared to the previous models. Climax is a foundation model leveraging extensive datasets from the 6th phase of the Coupled Models Intercomparison Project (CMIP6) global climate models. It can be fine-tuned to address various climate and weather tasks, including global/regional weather forecasting and future climate projection. 
ClimaX and the original ViT paper \cite{dosovitskiy2021image} showed a better and more robust solutions when increasing the sequence length using the full attention matrix on ERA5 and ImageNet~\cite{deng2009imagenet} dataset respectively. In fact, \cite{dosovitskiy2021image} suggested that scaling the compute is be a better predictor of the model quality rather than scaling the model size alone for ViTs.





\section{Sequence Length Impact and Bottlenecks} \label{sec:seq-vit}
In this section, we describe how the image resolution and the multiple channels can affect the sequence length and bottlenecks that are encountered with the current approaches. 


\subsection{Compute Cost to Train ViTs on Frontier}

To better understand the computation cost for training on large sequence lengths and estimate the difference by scaling the model size, we can calculate the ideal training time as a function of the number of FLOPs for the ERA5 dataset, i.e. 350K time-steps. For the following calculations we used the formulas as described by \cite{10.1145/3581784.3613215}. The total number of floating point operations (FLOPs) needed to train a ViT, depends on the parameter of the model (P) and the number of tokens (D), where for each element of a feed-forward weight matrix, there is a total of 6 FLOPs (1 multiply-add operation during the forward pass and 2 multiply-add operations during the backward pass) per input token:

\begin{equation}
\label{eqn:1}
T_{FLOPS} \sim 6 * P * D
\end{equation}

Also, the training time ($t$) depends on the measured percentage of the GPU's theoretical peak ($R_{peak}$), the ratio of tokens over the parameters ($\gamma = T/P$) and the number of devices N:

\begin{equation}
\label{eqn:2}
t \sim \frac{6 * \gamma P^{2}}{R_{peak}N}
\end{equation}

The training time from the above equation assumes that the model will converge on such a large batch size, the kernels libraries used by the ViT architecture can fully utilize the compute of the device, and the scaling efficiency is 100\% across Frontier (so we don't account at all internode and internode communication bandwidth, IO, algorithmic and libraries scaling inefficiencies).  

\begin{figure}[h!]
\centering
\hspace*{-.1cm}\includegraphics[width=.49\linewidth]{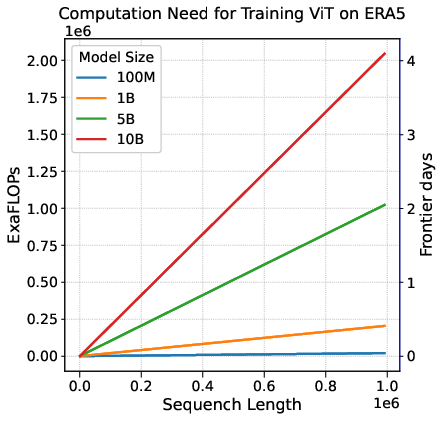}
\centering
\hspace*{0cm}\includegraphics[width=.49\linewidth]{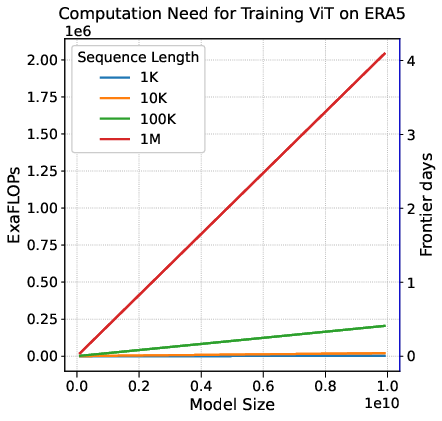}
\caption{Computation cost in FLOPs by varying the number of tokens (left plot), and the parameter size of the model (right plots). For both plots equations (\ref{eqn:1}), and (\ref{eqn:2}) were used with a total number of tokens for the ERA5 datasets, i.e. around 350K images}
\label{fig:cost_vit}
\end{figure}

Figure \ref{fig:cost_vit} was made using equations (\ref{eqn:1}), and (\ref{eqn:2}) by varying the number of tokens (left plot) and the model size (right plot). As we can see from the plots scaling the sequence length is as big of a problem as scaling the model size, and even though the two are orthogonal to each other, they are both going to add to the total computational cost and detailed study, as this, is necessary to strike a balance of the two and to find the optimal ratio between sequence-length and model size.

\subsection{Sequence Length for Scientific Images}
Scientific images are usually found in very high resolution on multiple channels. 
Data resolution in Earth system science can vary depending on the targeted phenomena and whether the data comes from simulations or satellites. In our current work, the primary dataset utilized for convergence runs was the ECMWF Reanalysis v5 (ERA5) \cite{ERA5}. ERA5 represents the fifth generation of reanalysis by the European Centre for Medium-Range Weather Forecasts (ECMWF), covering global climate and weather for the past eight decades. It provides hourly estimates across a range of variables that represent the atmosphere, ocean, and land surface. Typically, a ViT patch size of 2 is employed to embed the climate data, which, when trained on low resolution, results in an 8K sequence length \cite{nguyen2023climax}. The native resolution of ERA5 is 770x1440 grid points, leading to a 277K sequence, which is impractical to accommodate on a single GPU. Moreover, weather models, such as those referenced in \cite{HRRR}, can have resolutions up to 100 times greater than those of climate models in CMIP6. Additionally, these datasets may feature more than 100 channels, culminating in sequence lengths that extend into the millions.

\subsection{Training Results for Different Sequence Length}
Given the many ways the sequence length can be increased with climate data, along with the growing interest in developing Earth system foundational models based on ViTs, we think this application space is ideal for studying the effect of sequence length.


With the ERA5 dataset, we choose 5 input atmospheric variables: geopotential, temperature, u-component of wind, v-component of wind and specific humidity on 17 pressure levels: 10, 20, 30, 50, 70, 100, 150, 200, 250, 300, 400, 500, 600, 700, 850, 925, 1000, 3 surface level variables: 2m temperature, 10m u-component of wind, 10m v-component of wind, and 2 statics: orography and land-sea mask. Collectively, this combination leads to a total of 92 variables (or channels).  The performance is evaluated for 4 variables: 2m temperature (T2m), 10m u-component of wind (U10), geopotential at 500hPa (Z500), and temperature at 850hPa (T850). We also re-gridded the original dataset from ~0.25° (770 × 1440) to 3 low resolutions for the experiments: 1.0° (180 × 360), 1.40625° (128 × 256), and 5.625° (32 × 64). The re-gridding is done by utilizing the xESMF \cite{xesmf2020} based on the high-performance Earth System Modeling Framework (ESMF) as the backend, and it performs re-gridding between the general curvilinear grids with different re-gridding algorithms, e.g., bilinear, nearest neighbor and conservative, we applied the bilinear method to all of our re-gridding tasks.

\begin{figure}[h!]
\centering
\hspace*{-.5cm}\includegraphics[width=1.\linewidth]{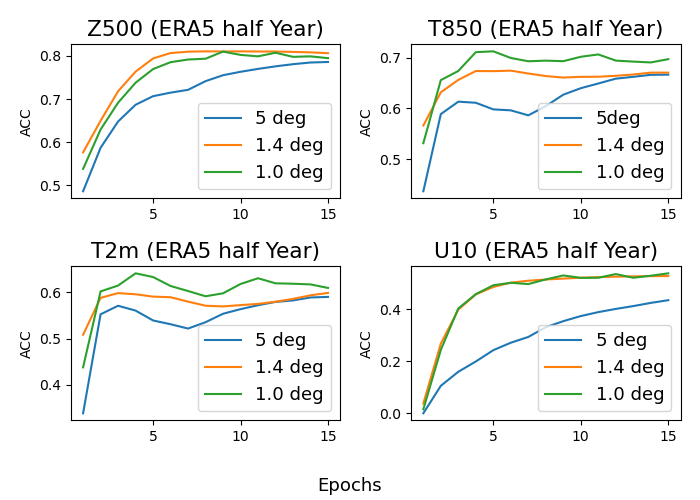}
\caption{Validation accuracy on half-year of the ERA5 dataset of the Z500, T850, T2m, and U10 variables. The effect in the accuracy is shown for three different spatial resolutions: 5.625°, 1.40625°, and  1.0°. A patch size of 4 was used, and so the sequence length for each resolution is 128, 2048, and 4050 respectively.}
\label{fig:acc_res_effect}
\end{figure}

\begin{figure}[h!]
\centering
\hspace*{-.5cm}\includegraphics[width=1.\linewidth]{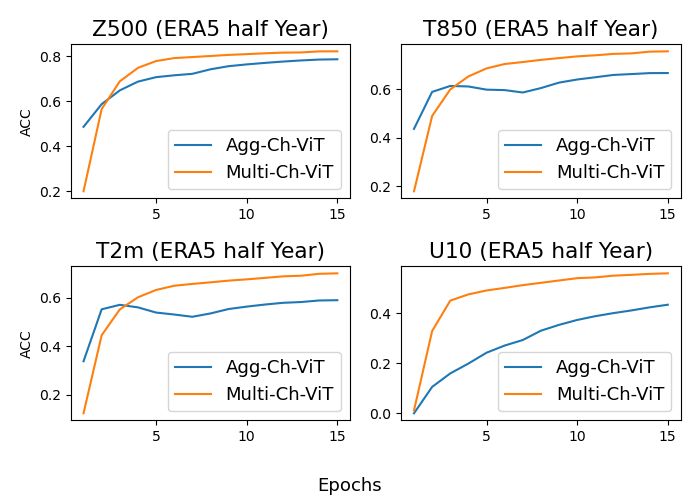}
\caption{Validation accuracy on half-year of the ERA5 dataset of the Z500, T850, T2m, and U10 variables. The effect in the accuracy is shown for including all 92 variables in the model, Multi-Ch-ViT, versus Agg-Ch-ViT. The sequence length of the first approach is 11,776, while it reduces to 128 for the second approach.}
\label{fig:acc_var_effect}
\end{figure}


We used the first half of the 2018 year for training and the second half for testing. All the hyper-parameters were kept constant between each experiment, using a predicted range of 28 hours, with a lead-time of 6 hours. Figure \ref{fig:acc_res_effect} shows the effect of three different spatial resolutions: 5.625°, 1.40625°, and  1.0°. Since we use a constant patch size of 4, it will result in the following sequence lengths: 128, 2048, and 4050, respectively. Figure \ref{fig:acc_var_effect} shows the effect of Multi-Ch-ViT, by using all 92 channels for each time-step, against the Agg-Ch-ViT as implemented in \cite{nguyen2023climax}. This will also increase the sequence length from 128 to 11,776 tokens per time step. 

Even though experiments in Figures \ref{fig:acc_res_effect} and \ref{fig:acc_var_effect} are different, they both show that increasing the sequence length results in a better overall accuracy. Also it is worth noting, increasing the resolution leads to a better accuracy from the very first epoch, while for the multi-ch-ViT, we see improvements for the majority of the predicted variables after the 4th epoch. The reason for the latter is in multi-ch-ViT the sequence length increases 92 times from the baseline, compare to 32 time for the resolution increase, and so it takes longer time for the attention layers to build a good set of weights, and learn the relationships between the 92 different channels. Another observation is that in both cases, the Z500 doesn't see as big of an improvement as the other variables. Insensitivity of Z500 is partly expected as it exhibits spatially less varying synoptic-scale features in the middle troposphere. Therefore, increasing the resolution does not add significantly distinct details. On the other hand, the T850 in the lower troposphere expectedly has more spatial variability due to the influence of the land-atmosphere flux exchanges. Hence, it shows improvements in results for the Multi-Ch-ViT compared to Z500.


From the experiments conducted thus far, we have identified a need to increase the sequence length for ViTs. Unstructured raw data, such as images and videos, can significantly escalate the number of tokens on which transformers are trained, presenting substantial computational challenges. The optimal balance between image resolution and the number of channels likely varies depending on the domain, necessitating a careful consideration of computational costs. For small models dealing with very high-resolution images, aggregating the channels may be the preferable strategy. Conversely, when dealing with a large number of channels but with moderate resolution, it is advisable not to aggregate the channels. Instead, increasing the patch size to manage the sequence length more effectively in relation to the image resolution is recommended.



\section{Scaling Solutions for Large Sequence Length and Evaluation} \label{sec:vit-perf}

In this section, we evaluate the performance implications of extending the sequence length using different distribution methods on Frontier's AMD GPU. Specifically, we incorporate two sequence parallelism strategies into ViTs: DeepSpeed-Ulysses and LSS, each adopting a significantly different approach to distributing the sequence length. We independently assess PP and TP, in addition to integrating them with the aforementioned sequence parallelism techniques. DeepSpeed-Ulysses and PP are implemented within the DeepSpeed framework\cite{rajbhandari2021zeroinfinity}, whereas LSS and TP utilize the Fully Sharded Data Parallel (FSDP) framework~\cite{zhao2023pytorch}. Given the variance in framework integration, a direct comparison between LSS and DeepSpeed-Ulysses is not straightforward; however, we can compare their scaling efficiency. Additionally, we offer a practical guide to determine the most suitable approach under varying conditions.


\subsection{Frontier Architecture}
All the experiments were performed using the Frontier Supercomputer \cite{Frontier} at the Oak Ridge Leadership Computing Facility. Each Frontier node has a single 64-core AMD EPYC CPU and four AMD Instinct MI250X GPU accelerators. The MI250X GPU is comprised of two Graphics Compute Dies (GCDs), connected with Infinity Fabric CPU-GPU, while the four MI250X GPUs are connected with Infinity Fabric GPU-GPU of 50GB/s. The system identifies each GCD independently, so from the application perspective, it can be considered that each node has 8 GPUs, each with 64 GB of high-bandwidth memory. For simplicity, we will use the term GPU when referring to a GCD. The nodes are connected via a Slingshot-11 interconnect with 100GB/s to a total of 9408 nodes, making it the first true exascale machine. For the software stack, we used  Pytorch 2.4 nightly build 03/16/2024. ROCm v5.7.0, MIOpen v2.19.0, RCCL v2.13.4 with libfabric v1.15.2 plugin.

\subsection{Flash Attention v2 on Frontier}
\begin{figure}[h!]
\centering
  \centering
  \hspace*{-1cm}\includegraphics[width=1\linewidth]{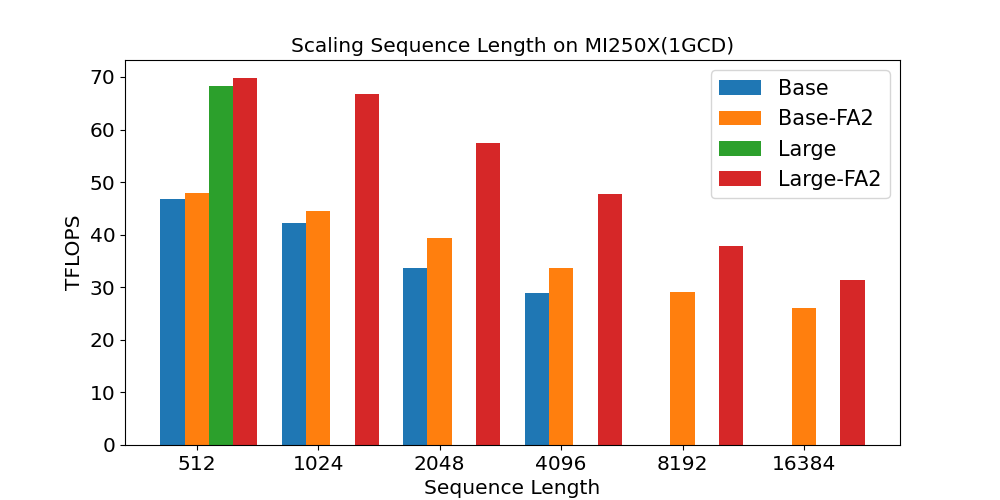}
  \centering
  \hspace*{-1cm}\includegraphics[width=1\linewidth]{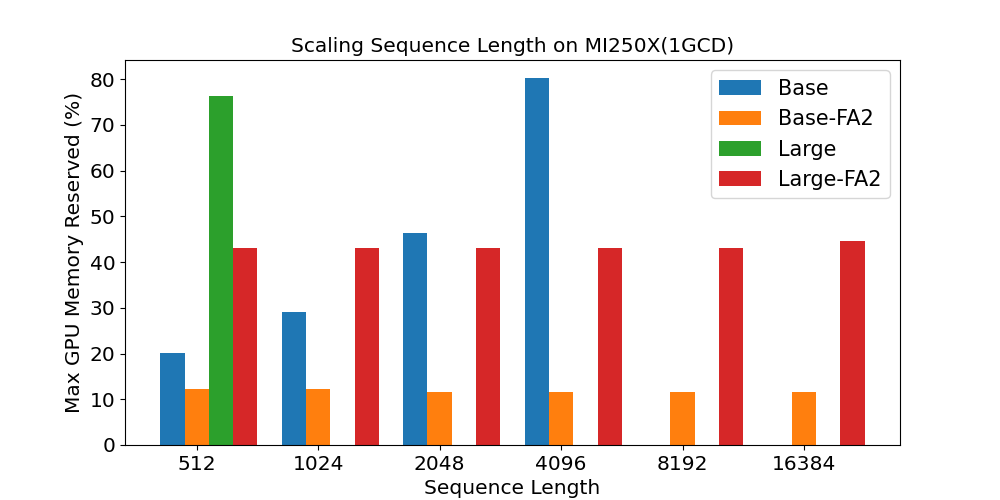}
\caption{Single GPU measurements with and without flash-attention-2, on Frontier MI250X GPU (1 GCD). The top plot shows the GPU throughput for two model sizes, ViT-Base and ViT-Large, while the bottom plot shows the max GPU memory reserved throughout each run. For the flash-attention-2 measurements, Pytorch 2.4 nightly build 03/16/2024 was used with rocm5.7.}
\label{fig:fa2}
\end{figure}

Figure \ref{fig:fa2} shows the compute performance in FLOPs and the memory footprint as a function of the sequence length, for Flash-Attention-v2 (FA2) \cite{dao2023flashattention2} on Frontier's AMD MI250X GPU. For the base model, a hidden dimension of 1024 was used with 16 attention heads, while for the large model, a 2048 hidden dimension was used with 32 attention heads. We scale the sequence length from 512, to 16K, and set the batch size so that the total number of tokens is 16K through each run. 

The first main observation is that we can fit 4 times larger sequence lengths using FA2 for the base model and 32 times larger for the large model, which is a significant improvement when scaling the sequence length. For the base model, we also see improvements in the number of computational efficiencies, especially as we move to a larger sequence length. Also, the max reserved memory is reduced by 50\% or more in all cases. 

Overall, we don't observe big performance improvements using FA2 regarding computational efficiency, but we see savings in memory footprints. It is worth noting that the small performance gain doesn't reflect the AMD hardware itself but rather the readiness of the software implementation of FA2 on the ROCm software stack. For example, on a GPT model, \cite{yin2024comparative} has shown a performance gain of 20\% on the Frontier MI250X GPUs using the open-sourced development \cite{amd-fa2} of flash attention v1. In this work, we directly used Pytorch fused attention kernels, while we didn't write from scratch rocm code for our specific model architecture. 

Given that FA2 performance is still in the experimental stage within Pytorch on AMD GPUs, we will not use it when searching for the best-performing distributed approach but only after finding the best parallel strategy to push the highest sequence length per GPU.

\begin{figure}[h!]
\centering
  \centering
  \hspace*{-.5cm}\includegraphics[width=1.\linewidth]{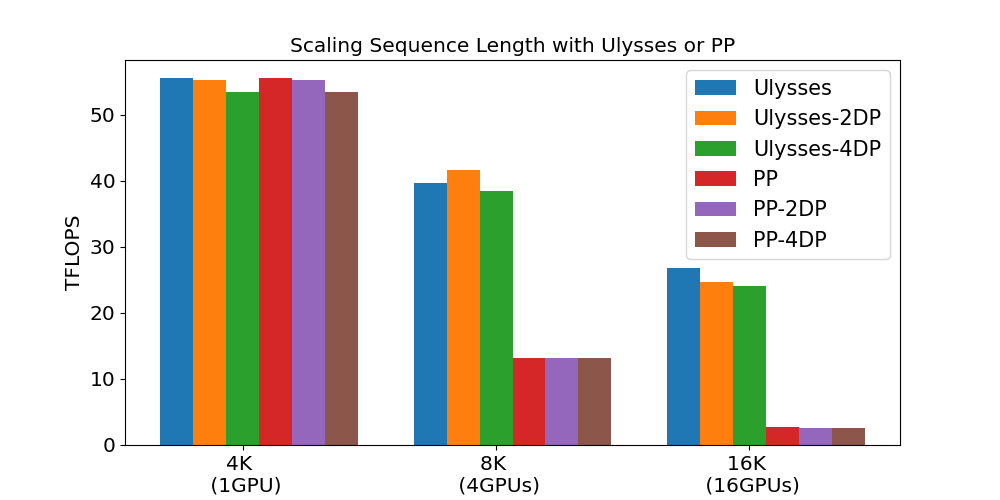}
  \centering
  \hspace*{-.5cm}\includegraphics[width=1.\linewidth]{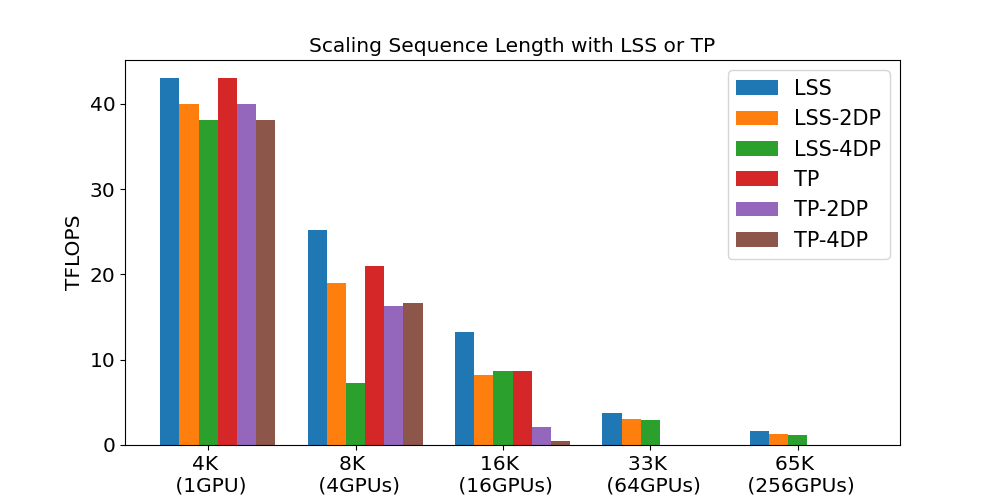}
\caption{Weak scale the sequence length for each distributed method separately. Specifically, the top plot shows the DeepSpeed-Ulysses and Pipeline (PP) methods within the DeepSpeed framework, while we also scale the batch size linearly with the DeepSpeed Zero Redundancy Optimizer (DP). The bottom plot shows the Long Short-Sequence method (LSS) and the tensor parallelism (TP) methods within Pytorch's FSDP framework, while we also show scaling linearly the batch size with the fully sharded method (DP). The model for the two implementations was set to a 2560 hidden dimension with 16 heads total.}
\label{fig:seq_scale_separate}
\end{figure}

\subsection{Different Distributed Approaches}
We use DeepSpeed-Ulysses, PP, LSS, and TP separately to scale the sequence length beyond the single GPU limits. The top plot of Figure \ref{fig:seq_scale_separate} shows DeepSpeed-Ulysses and PP used separately by weak scaling the sequence length within the DeepSpeed framework. The bottom plot of Figure \ref{fig:seq_scale_separate} shows LSS and TP used separately by weak scaling the sequence length within Pytorch's FSDP framework. In all four cases, we also measured the performance of scaling the batch for each method separately using data-parallel approach. A ViT model architecture of 2560 hidden dimensions was used, with 16 attention heads in total.

As we can see DeepSpeed-Ulysses shows the best scaling efficiency from 1 to 16 GPUs. Next, we see that LSS scales better than TP and PP. That behavior was expected since PP and TP are designed to distribute the model size rather than the sequence length. We also observe that the DeepSpeed data-parallel scale is better for DeepSpeed-Ulysses and PP, compared to FSDP model sharding for LSS and TP. Probably, that's an indication that some tuning is necessary between FSDP and LSS. Finally, even though DeepSpeed-Ulysses scales better than LSS, there is a fundamental limit in the number of sequence parallel ranks that we can use, which is imposed by the number of heads. For example, the model architecture we used had 16 attention heads, so we can't use more than 16 sequence parallel ranks, limiting the sequence length we can run to 16K. 


\begin{figure}[h!]
\centering
\hspace*{0cm}\includegraphics[width=0.8\linewidth]{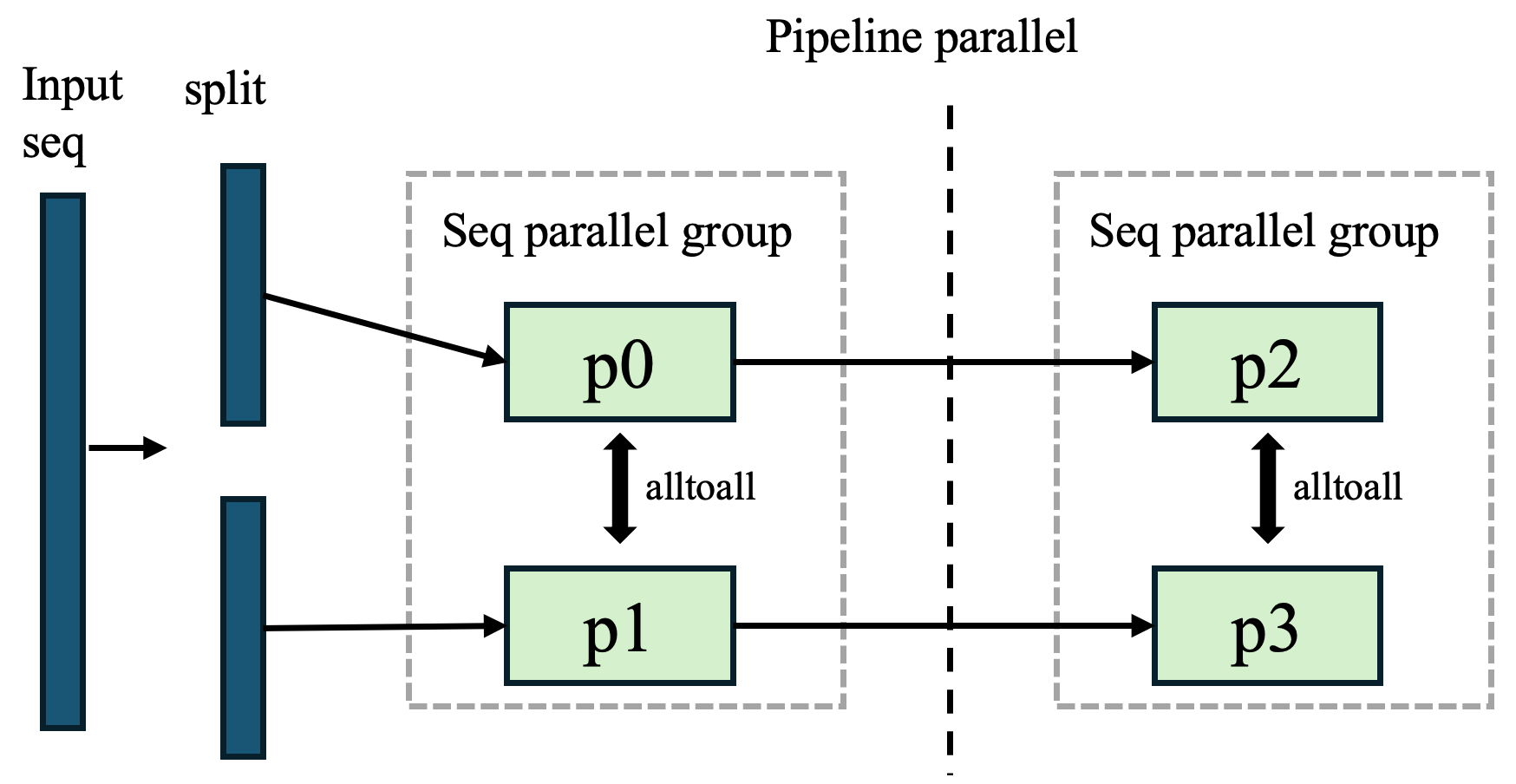}
\caption{The overview of combining sequence parallelism and pipeline parallelism.}
\label{fig:dist_PP_Uly_diagram}
\end{figure}

\begin{figure}[h!]
\centering
  \centering
  \hspace*{-.5cm}\includegraphics[width=.52 \linewidth]{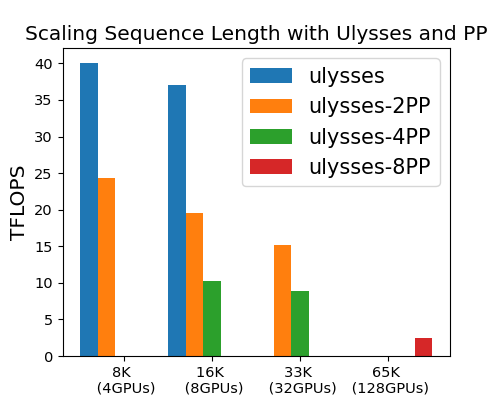}
  \centering
  \hspace*{0cm}\includegraphics[width=.52 \linewidth]{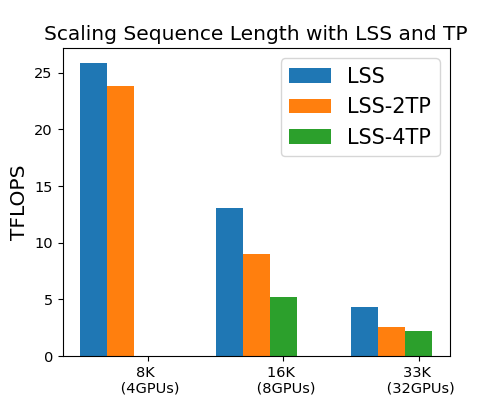}
  \centering
  \hspace*{-.5cm}\includegraphics[width=.52 \linewidth]{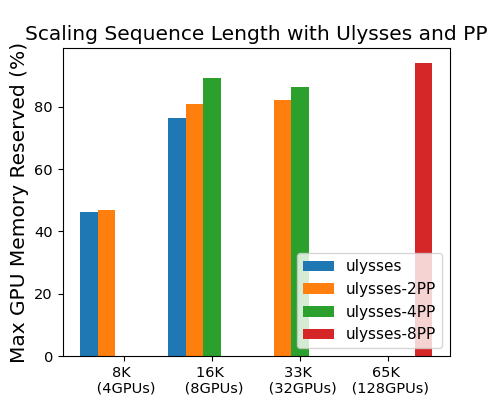}
  \centering
  \hspace*{0cm}\includegraphics[width=.52 \linewidth]{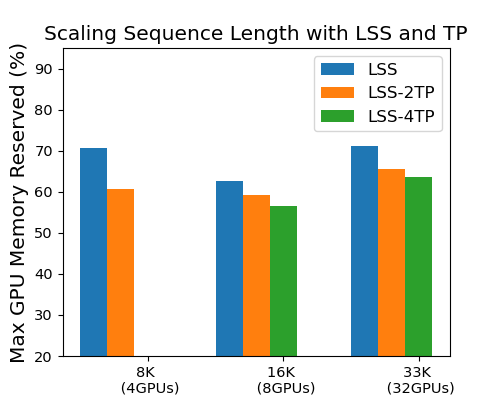}  
\caption{Weak scale the sequence length by pairing the DeepSpeed-Ulysses with Pipeline parallelism (PP) and the Long Short-Sequence method (LSS) with Tensor Parallelism (TP). The top plots show the GPU throughput, while the bottom plots show the max GPU memory reserved throughout each run. The number of PP (TP), shows how many pipelines (tensor parallel) ranks were used, and the DeepSpeed-Ulysses (LSS) ranks were adjusted to keep the number of GPUs constant. For example, for 8 GPUs, the Ulysses-2PP means that we did sequence parallelism within 4 GPUs, and between each GPU pair, we did pipeline parallelism, while the Ulysses-4PP means that we did pipeline parallelism within 4 GPUs, and between each GPU pair, we did sequence parallelism.}
\label{fig:seq_scale_together}
\end{figure}

\subsection{Hybrid Distributed Approaches}
Here, we present the results of combining DeepSpeed-Ulysses with Pipeline Parallelism (PP) and Long-Short Sequence (LSS) with Tensor Parallelism (TP) to scale the sequence length.

Often, it is beneficial to integrate distributed methods to optimize their advantages and mitigate potential bottlenecks. This approach is justified because: (1) Sequence parallelism, TP, and PP are orthogonal to each other. (2) Communication overhead can significantly restrict scalability if solely relying on sequence parallelism, TP, or PP. To address this, we combined LSS with TP to reduce the communication overhead involved in distributing activations from sequence groups to tensor parallel groups. Furthermore, we integrated DeepSpeed-Ulysses with PP (see an example configuration in Figure~\ref{fig:dist_PP_Uly_diagram}). 
DeepSpeed-Ulysses primarily utilizes two all-to-all communication calls in both the forward and backward passes, complemented by an all-reduce per layer, and concludes with an all-gather call at the end of each step for optimization. In contrast, PP mainly involves all-reduce operations between partitioned layers.

Additionally, the top plot in Figure \ref{fig:seq_scale_together} illustrates the compute throughput in TFLOPs for pairing two methods, whereas the bottom plot depicts the memory footprint of the application. The labels 2PP, 4PP, and 8PP indicate the number of pipeline ranks, with the rest of the configurations dedicated to distributing the sequence length. For instance, "Ulysses-2PP" on 8 GPUs signifies that there were four groups employing sequence parallelism with DeepSpeed-Ulysses, each consisting of two GPU pairs engaged in pipeline parallelism. This rationale is similarly applied to LSS-2TP and other configurations.

Our first observation is that we successfully overcame the limitation of DeepSpeed-Ulysses, which was only capable of handling a number of attention heads equal to the distributed ranks. We are now able to scale beyond a 16K sequence length, thanks to the ability to utilize additional ranks with pipeline parallelism, scaling up to 65K on 128 GPUs. This is achieved by employing 8 GPUs per node for pipeline parallelism and distributing the sequence length across nodes. Using DeepSpeed-Ulysses alone in the previous subsection, we observed substantial scaling benefits as the ranks increased. However, integrating it further, we now witness a significant performance enhancement. We attribute the primary reason for this improvement to communication dynamics; as indicated by the bottom left plot, our memory footprint exceeds 95\%, suggesting that memory consumption increases with more ranks, contrary to our experience with solely DeepSpeed-Ulysses.

On the other hand, LSS is not confined by the number of sequence ranks deployable, yet its scaling efficiency did not match that of DeepSpeed-Ulysses. Thus, in the right plot of Figure \ref{fig:seq_scale_together}, we experimented with combining LSS with Tensor Parallelism (TP) to enhance compute performance. The memory footprint data reveal that, despite employing additional ranks with TP, we were unable to reduce the memory usage below 50\% of the maximum, which would allow for an increase in the local batch size and overall computation. Interestingly, a comparison between the memory footprints of DeepSpeed-Ulysses with Pipeline Parallelism (PP) versus LSS with TP shows an increase in memory usage in the former case and a decrease in the latter. This outcome is promising for LSS and TP configurations, as it suggests potential for further optimization to accommodate larger local batch sizes, thereby improving compute throughput.

\subsection{Combining DeepSpeed-Ulysses with FA2}
In this subsection, we use the best performing distributed approach we found so far, i.e., DeepSpeed-Ulysses with Flash-attention-v2 to push the limits of the sequence length on Frontier with our current software stuck. For our tests, we used models with four different sizes, as defined in Table \ref{table:models}. 

 \begin{table}[h!]
 \caption{A Summary of the different model \\ architectures explored in this work}
\centering
\begin{tabular}{ |c|c|c|c|c|c| }
 \hline
 \textbf{Model} & \textbf{Width}  & \textbf{Depth} & \textbf{Heads} & \textbf{Params [M]} \\
 \hline
 ViT-Base &   1024  & 8  &  16  & 84 \\
 ViT-Large &   2048  & 16  &  32  &  671 \\
 ViT-5B &   4096  & 32  &  32  & 5,370 \\
 ViT-10B &   4608  & 48  & 64  & 10,194 \\
 \hline
\end{tabular}
\label{table:models}
\end{table}

\begin{figure*}[h!]
\centering
  \centering
  \hspace*{0.cm}\includegraphics[width=.34\linewidth]{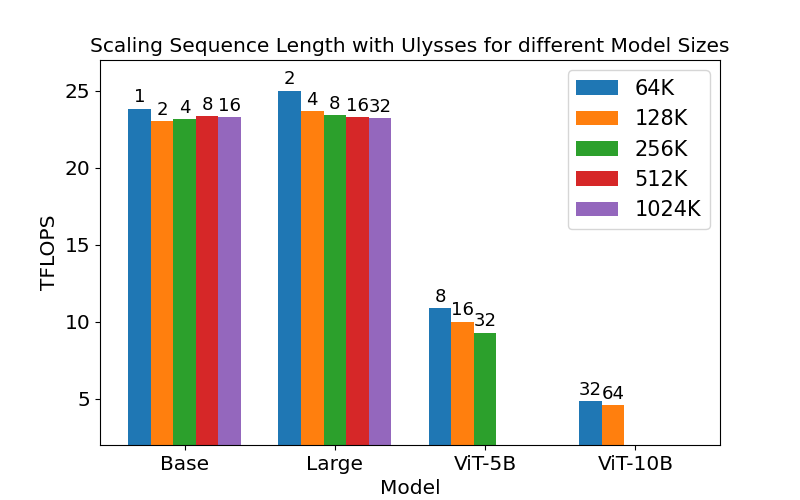}
  \centering
  \hspace*{-0.5cm}\includegraphics[width=.34\linewidth]{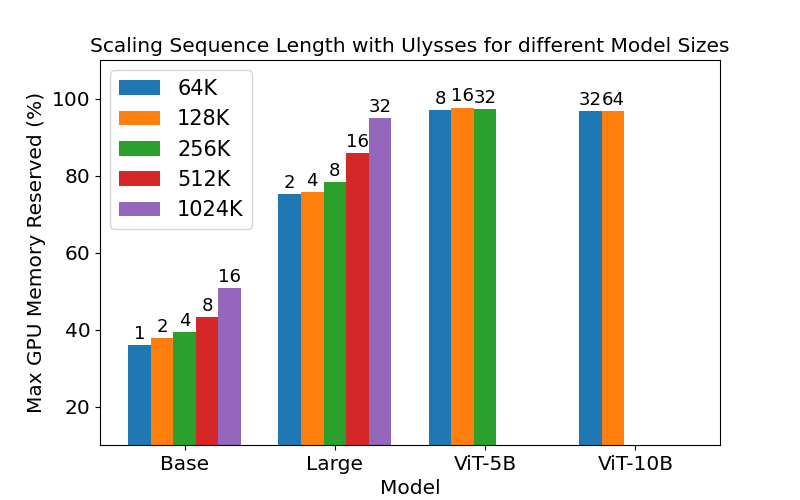}
  \centering
  \hspace*{-.5cm}\includegraphics[width=.34\linewidth]{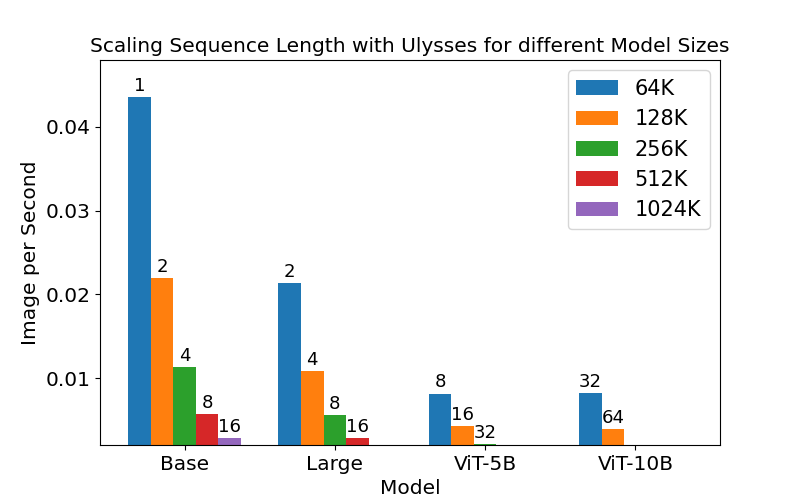}  
\caption{Measuring the GPU throughput, the memory footprint, and the image-per-second for four model sizes (see Table \ref{table:models}), by scaling the sequence length. Only the DeepSpeed-Ulysses distributed method was used, along with DeepSpeed Zero Redundancy Optimizer and flash-attention-2. The numbers on top of the bar charts show the number of GPUs for each measurement.}
\label{fig:scale_Ulysses_FA2}
\end{figure*}


Figure \ref{fig:scale_Ulysses_FA2} shows the compute throughput, the memory footprint, and the image-per-second for the four different models for different sequence lengths. The ViT-Base and the ViT-Large model can scale up to 1M sequence length on 16 and 32 GPUs respectively. The compute scaling efficiency looks very good in both cases, which indicates that we are not compute bound for those models, even up to 1M sequence length. On the other hand, we can see the memory increases almost linearly for the base and large model, mostly an expected behavior since we are using flash-attention on the compute, and communication doesn't seem to be affecting the performance. Finally, we can see for the base and large model, 16 times less image-per-second performance from 64K to 1M sequence length. This is also expected since we are increasing the sequence length, and so the number of tokens per training step, almost by the same amount, going from 64K to 1M.

Figure \ref{fig:scale_Ulysses_FA2} also shows the performance of the ViT-5B and ViT-10B models. To accommodate a 64K sequence length, a minimum of 8 GPUs is necessary for the ViT-5B, and a minimum of 32 GPUs is required for the ViT-10B. However, for both models, scaling up to a 1M sequence length using FA2 and DeepSpeed-Ulysses alone is not feasible. This limitation arises because, within DeepSpeed-Ulysses the number of sequence ranks cannot exceed the number of attention heads. Consequently, the maximum number of sequence ranks for the ViT-5B is 32, and for the ViT-10B, it is 64. We can overcome this limitation by integrating pipeline parallelism as shown before.

Finally, we can see from Figure \ref{fig:scale_Ulysses_FA2} that the ViT-5B and ViT-10B models show poorer compute performance compared to base and large models. Both of those models won't be able to fit on a single GPU, so we are relying on DeepSpeed to share the parameters and activations of the model. This will increase the overall communication of the application since apart from the all-to-all communication that DeepSpeed-Ulysses is primarily using, we are going to have additional all-gather and scatter due to the larger model size.  We can also see that the memory footprint of the two larger models is constant as we scale the sequence length, which is an indication that the application at that scale doesn't have ideal compute and communication overlap and some further fine-tuning might be necessary. 

\subsection{Scaling Sequence Length}
Based on these experiments, we found that if we want to use a single distributed strategy to scale the sequence length for smaller models the best performing approach is DeepSpeed-Ulysses along with FA2. Also combining DeepSpeed-Ulysses with DeepSpeed Zero parallel modes shows better out-of-the-box scaling performance compared to comping FSDP with other distributed strategies. For models that can't fit on a single GPU DeepSpeed-Ulysses alone is limited, and an overlap between PP and DeepSpeed-Ulysses is necessary. The latter will probably have its upper bound limit since the memory footprint keeps increasing, while we increase the number of ranks, even by keeping the local batch size constant. On the other hand, even though the LSS method does show less scaling efficiency, compared to DeepSpeed-Ulysses, it doesn't suffer from a limit in the number of sequence ranks to use. Furthermore, when combined with TP, since the memory footprint does decrease with more ranks, it shows more potential to fit larger local batch size, and so at a very large scale of model and sequence length probably a combination of LSS and TP would be the best approach. 

\section{Training Long Sequence ViTs for Climate Data} \label{sec:vit-all}
Here, we study the performance and training results on climate data, using FA2 and DeepSpeed-Ulysses on two model sizes, ViT-Base and ViT-Large (as defined in Table~\ref{table:models}) for up to 188,416 sequence length. We first show the performance of weak scaling the batch size of a model with 188,416 sequence length up to 2048 GPUs on Frontier.

\subsection{Performance at Scale}
\begin{figure}[h!]
\centering
  \centering
  \hspace*{-.5cm}\includegraphics[width=1.\linewidth]{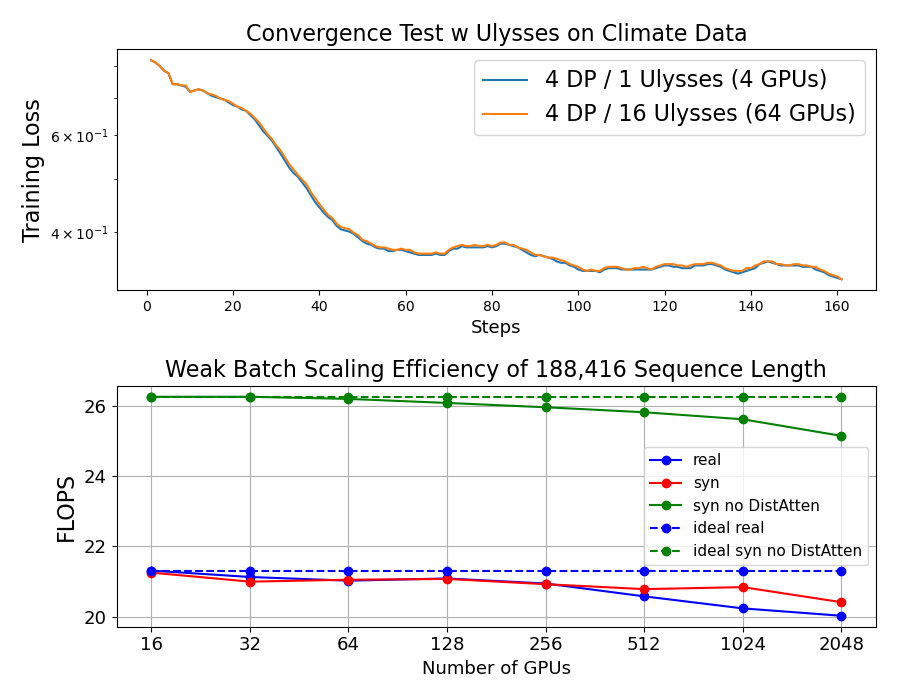}
\caption{The top plot shows a convergence comparison between a single GPU sequence length and distributing the sequence to 16 GPUs using DeepSpeed-Ulysses. Real ERA5 data was used for this test, along with scaling the batch size to 4 with DeepSpeed Zero Optimizer in both cases. In the bottom plot we weak scale the batch size of a 188,416 sequence length base model up to 2,048 GPUs. The real plot is with real ERA5 data, the synthetic plot is with random data, and the "no DistAttent" is with synthetic data but turning off distributed attention. A minimum of 16 GPUs was necessary to fit the base model with a 188,416 sequence length with a local batch size of 4 that was used for the convergence runs.}
\label{fig:weak_ulysses_fa2}
\end{figure}

First, we think it is essential when a distributed method, beyond data parallel, is integrated into a model to showcase numerical correctness when converging to a solution on real data. To that extent, the top plot of Figure \ref{fig:weak_ulysses_fa2} compares the same model and sequence length for 4 DeepSpeed Zero ranks against 16 sequence parallel ranks with DeepSpeed-Ulysses along 4 DeepSpeed Zero ranks in each group. We can see that for the first steps the loss is identical, as expected, and we further test that in the first forward pass, all matrices are the same. We then see some small differences as training progresses, which primarily come from the non-deterministic parts of the model, such as normalization layers and dropout. Additionally, we see that the model and the data are seeded correctly as we are performing the comparison experiments next.

\begin{figure}[h!]
\centering
  \centering
  \hspace*{-.5cm}\includegraphics[width=1.\linewidth]{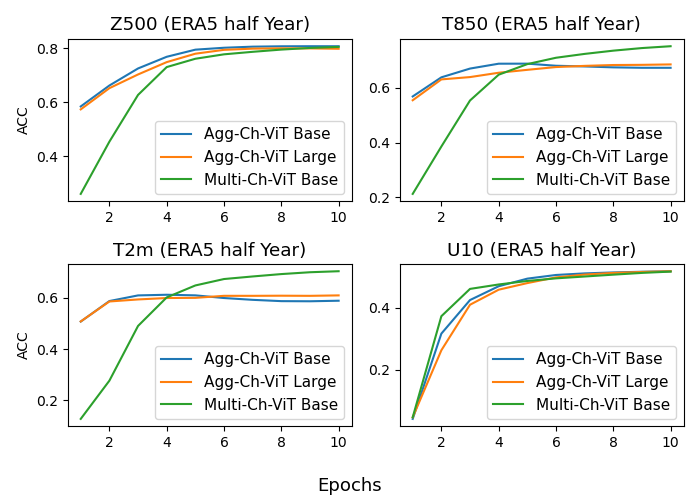}
\caption{The plot shows the validation accuracy on half-year of the ERA5 dataset of the Z500, T850, T2m, and the U10 variables. The effect in the accuracy is shown between a base Agg-Ch-ViT model with 2,048 sequence length (aggregate 92 variables to 1), a large Agg-Ch-ViT model with 2,048 sequence length (aggregate 92 variables to 1), and a base Multi-Ch-ViT model with 188,416 sequence length (all 92 variables of the ERA5 dataset were used).}
\label{fig:acc_ulysses_fa2}
\end{figure}

The bottom plot of Figure \ref{fig:weak_ulysses_fa2} shows a weak scaling of the batch size for the 188,416 sequence length model on Frontier. In the training, we found a better convergence with a local batch size of 4, and for this, we need a minimum of 16 GPUs to fit the 188K sequence length for the ViT-Base model. With this as the baseline, using DeepSpeed Zero, we scale up to 2,048 GPUs. The blue curve runs are with real climate data, showing a scaling efficiency of 94\%. The red curve shows a 96\% scaling efficiency with synthetic data, confirming that the application isn't IO-bound as expected. The green curve also represents a run with synthetic data, but this time, we don't wrap the attention layers with DeepSpeed-Ulysses. This would be a way to estimate the communication overhead from distributing the sequence length, which, as we mentioned before, is mostly two all-to-all calls in the forward pass and two all-to-all calls + all reduce in the backward pass per layer. Since we get the same scaling efficiency as in the red curve, that probably means that the communication part of DeepSpeed-Ulysses scales very well with the batch size, but the constant 20\% reduction in the number of FLOPS that we see probably relates to the overlap between compute and communication within the sequence parallel groups.


Second, we show the training results using the ERA5 dataset, with the same setup as in Section \ref{sec:seq-vit}. trained with a resolution of 1.40625°. All the hyper-parameters were kept constant between each experiment, using a predicted range of 28 hours, with a lead-time of 6 hours. Three experiments are shown: one with Agg-Ch-ViT and the ViT-Base model, one with Agg-Ch-ViT and the ViT-Large model, and one with Multi-Ch-ViT and the ViT-Base model. The sequence length for the first two experiments is the same, 2,048 but the model size is increased from 100M to 800M, while in the latter experiment, the sequence length is 188,416.

Figure \ref{fig:acc_ulysses_fa2} shows results from all three experiments. With larger sequence lengths, Z500 and U10 don't show improvements, while T850 and T2m seem to show better accuracy. The advantage of the Multi-Ch-ViT is that we explicitly pass the model to all the channels per time-step, compared to the Agg-Ch-ViT, where all channels are aggregated into one. As discussed in Section \ref{sec:seq-vit} for the Z500, it is expected to not see much improvement since Z500 features won't have fast spatiotemporal variations. In comparison, the temperature on the surface and lower troposphere exhibits substantially more variations. Also, we observed the same behavior as in Section \ref{sec:seq-vit}, where the convergence starts at lower performance, and with more epochs, the model can utilize better the longer sequence length. In principle, if the model could understand the full climate phenomena, this should reflect on the better performance of the Z500 as well, but the number of epochs was not as large to see this effect. Finally, we see that for those experiments, increasing the model size didn't result in a better solution. Probably 8 times larger model size isn't enough to see any major improvements, but because we only used 1 year of data for those experiments, we can't justify a larger model. 


Finally, we show that Ulysses also scales well with the batch size, using DeepSpeed Zero as the data-parallel mode, while IO can successfully hide behind compute even for 188K sequence length. Even though we only trained for 10 epochs, we still observe almost 20\% improvements in the accuracy for the temperature channels, while we recommend as the sequence length becomes larger, a longer number of epochs is necessary. Also for ViTs, a practitioner should consider scaling the sequence length first, rather than the model size, though a balance between the two would be the optimal approach.

\section{Conclusion and Future Work}

We developed a ViT capable of processing ultra-long sequence lengths of up to 1 million tokens, employing sequence parallelism with DeepSpeed-Ulysses and LSS strategies. This approach, tested across models with up to 10 billion parameters, addressed performance bottlenecks and utilized distributed sequence strategies alongside Pipeline and Tensor Parallelism to scale beyond single GPU memory capacities. Notably, applying these methodologies to climate modeling increased temperature prediction accuracy by up to 20\%, marking a significant advancement in handling extensive sequence lengths with full attention.

Future research will focus on refining parallelism strategies for enhanced efficiency and exploring the application of ultra-long sequence processing across various domains. Investigating alternative attention mechanisms to optimize computational demands while preserving model performance will also be a key area of interest, potentially broadening the utility and scalability of ViTs in data-intensive fields.

\section*{Acknowledgments}
This manuscript has been authored by UT-Battelle, LLC, under contract DE-AC05-00OR22725 with the US Department of Energy (DOE). The US government retains and the publisher, by accepting the article for publication, acknowledges that the US government retains a nonexclusive, paid-up, irrevocable, worldwide license to publish or reproduce the published form of this manuscript, or allow others to do so, for US government purposes. DOE will provide public access to these results of federally sponsored research in accordance with the DOE Public Access Plan (http://energy.gov/downloads/doe-public-access-plan). This research used resources of the Oak Ridge Leadership Computing Facility, which is a DOE Office of Science User Facility supported under Contract DE-AC05-00OR22725.

\bibliographystyle{unsrt} 
\bibliography{sample-base}

\end{document}